\documentclass{article}
\usepackage{arxiv}

\usepackage[utf8]{inputenc} 
\usepackage[T1]{fontenc}    
\usepackage{hyperref}       
\usepackage{url}            
\usepackage{booktabs}       
\usepackage{amsfonts}       
\usepackage{nicefrac}       
\usepackage{microtype}      
\usepackage{lipsum}
\usepackage{graphicx}

\usepackage{amsmath,amssymb} 
\usepackage{color}

\usepackage{times}
\usepackage{latexsym}
\usepackage{lmodern,blindtext}
\usepackage{microtype}

\usepackage{amsmath}
\usepackage{amsfonts}
\usepackage{amssymb}
\usepackage{wrapfig}
\usepackage{subcaption}
\usepackage{multirow}
 \usepackage{mathtools} 
\usepackage{verbatim}
\usepackage{anyfontsize}
\usepackage{color}
\newcommand{\distas}[1]{\mathbin{\overset{#1}{\kern\z@\sim}}}%
\usepackage{enumitem}
\usepackage[lined,boxed,commentsnumbered,ruled,linesnumbered]{algorithm2e}
\usepackage{booktabs} 
\usepackage{colortbl} 
\definecolor{Gray}{gray}{0.93}

\newcommand{\RN}[1]{%
	\textup{\lowercase\expandafter{\it \romannumeral#1}}%
}

\newcommand{\beq}{\vspace{0mm}\begin{equation}}
\newcommand{\eeq}{\vspace{0mm}\end{equation}}
\newcommand{\beqs}{\vspace{0mm}\begin{eqnarray}}
\newcommand{\eeqs}{\vspace{0mm}\end{eqnarray}}
\newcommand{\barr}{\begin{array}}
\newcommand{\earr}{\end{array}}

\newcommand{\Kmat}[0]{{{\bf K}}\xspace}

\newcommand{\Qmat}[0]{{{\bf Q}}\xspace}
\newcommand{\Rmat}[0]{{{\bf R}}}

\newcommand{\Vmat}[0]{{{\bf V}}}

\newcommand{\vv}{\boldsymbol{v}}
\newcommand{\wv}{\boldsymbol{w}}
\newcommand{\xv}{\boldsymbol{x}}

\newcommand{\Lcal}{\mathcal{L}}

\DeclareMathOperator{\RR}{\mathbb{R}} 

\newcommand{\modelshort}{XGPT}
\newcommand{\modeltitle}{Cross-modal Generative Pre-Training for Image Captioning}
\newcommand{\modellong}{Cross-modal Generative Pre-Training for Image Captioning}
\newcommand{\modellongbf}{\textbf{Cross}-modal \textbf{G}enerative \textbf{P}re-\textbf{T}raining for Image Captioning}

\newcommand{\taskone}{Image-conditioned Masked Language Modeling}
\newcommand{\tasktwo}{Image-conditioned Denoising Autoencoding}
\newcommand{\taskthree}{Text-conditioned Image Feature Generation}

\newcommand{\taskoneshort}{IMLM}
\newcommand{\tasktwoshort}{IDA}
\newcommand{\taskthreeshort}{TIFG}

\newcommand{\lpretrain}{\textbf{Text Pre-training}}  
\newcommand{\vlpretrain}{\textbf{Image-Text Pre-training}}
\newcommand{\enconly}{Enc}  
\newcommand{\encdec}{EncDec}
\newcommand{\encdecshare}{EncDecShare}

\begin{document}

\title{\modelshort: \modeltitle}

\author{Qiaolin Xia\textsuperscript{$\clubsuit$}\quad Haoyang Huang\textsuperscript{$\diamondsuit$}\quad Nan Duan\textsuperscript{$\diamondsuit$}\quad 
Dongdong Zhang\textsuperscript{$\diamondsuit$}\quad
Lei Ji\textsuperscript{$\diamondsuit$}\quad\\
\textbf{Zhifang Sui}\textsuperscript{$\clubsuit$}\quad 
\textbf{Edward Cui}\textsuperscript{$\heartsuit$}\quad 
\textbf{Taroon Bharti}\textsuperscript{$\heartsuit$}\quad 
\textbf{Xin Liu}\textsuperscript{$\heartsuit$}\quad 
\textbf{Ming Zhou}\textsuperscript{$\diamondsuit$}\quad \\
\textsuperscript{$\clubsuit$}MOE Key Laboratory of Computational Linguistics, Peking University\quad\\ \textsuperscript{$\diamondsuit$}Microsoft Research Asia\quad \textsuperscript{$\heartsuit$}Microsoft\quad\\
{\tt \{xql,szf\}@pku.edu.cn}\quad
{\tt \{haohua,nanduan\}@microsoft.com}\quad\\
{\tt \{dongdong.zhang,leiji,edwac,tbharti,mingzhou\}@microsoft.com}
}
\maketitle
\begin{abstract}
While many BERT-based cross-modal pre-trained models produce excellent results on downstream understanding tasks like image-text retrieval and VQA, they cannot be applied to generation tasks directly.  
In this paper, we propose \modelshort, a new method of \modellongbf\ that is designed to pre-train text-to-image caption generators through three novel generation tasks, including \taskone\ (\taskoneshort), \tasktwo\ (\tasktwoshort), and \taskthree\ (\taskthreeshort). As a result, the pre-trained \modelshort\ can be fine-tuned without any task-specific architecture modifications to create state-of-the-art models for image captioning.
Experiments show that \modelshort\ obtains new state-of-the-art results on the benchmark datasets, including 
COCO Captions and 
Flickr30k Captions. 
We also use \modelshort\ to generate new image captions as data augmentation for the image retrieval task and achieve significant 
improvement on all recall metrics. 
\end{abstract}

\section{Introduction}
\label{sec:intro}
 
Cross-modal pre-training has substantially advanced the state of the art across a variety of Vision-and-Language (VL) tasks. VL understanding tasks, such as Image-Text Retrieval \cite{karpathy2015deep}, Visual Question Answering (VQA) \cite{antol2015vqa}, Visual Commonsense Reasoning (VCR) \cite{zellers2019VCR}, Referring Expression Comprehension \cite{kazemzadeh2014refexp}, require the pre-trained model to learn the representation of visual contents, language semantics, and cross-modal alignments, but don't require generation ability. Recent pre-training methods for understanding tasks \cite{lu2019vilbert,alberti2019b2t2,tan2019lxmert,li2019visualbert,li2019unicodervl,su2019vl,chen2019uniter} have achieved state-of-the-art performance 
and attracted a lot of attention in both CV and NLP. 

Vision-and-language generation tasks (e.g., Image Captioning and Text-to-Image Generation), however, requires the model to not only understand cross-modal representations but also learn generation capabilities. Thus, directly applying a model pre-trained on VL understanding tasks is not feasible. The reason is two-fold. On one hand, pre-trained models developed for understanding tasks only provides the encoder. To support generation tasks, separate decoders have to be trained, like the methods proposed by VideoBERT \cite{sun2019videobert} and CBT \cite{sun2019cbt}. On the other hand, existing VL pre-training objectives are almost all related to the masked region or span prediction, including VLP \cite{zhou2019unified}. None of the pre-training tasks is designed for the whole sentence generation. Compared to studies for understanding tasks, the large-scale pretraining and fine-tuning model for VL generation tasks are still under-developed.



In this paper, we present \modellong\ (\modelshort). The \modelshort\ model uses a cross-modal encoder-decoder architecture 
and is directly optimized for VL generation tasks. Inspired by UniLM \cite{dong2019unified},  we share parameters between the encoder and decoder to allow more effective cross-task knowledge sharing. 
To leverage underlying semantic and improve bidirectional generalization between two modalities, we carefully design three generative pre-training tasks: 1) \textit{\taskone} (\taskoneshort), 2) \textit{\tasktwo} (\tasktwoshort), 3) \textit{\taskthree} (\taskthreeshort), where the encoder takes single-stream or multi-stream data as input, and the decoder is adapted to predict a wide range of generation objectives in an autoregressive manner, such as words, a sentence or image features.




Following VisualBERT \cite{li2019visualbert}, we perform two-stage pre-training before fine-tuning, which allows the model to better adapt to the target domain. We propose several model variants and task combinations in a thorough ablation study, which allows us to carefully control a number of factors, including model architectures, training objectives, and whether to keep placeholders for span masks. 

In addition to vision-to-language generation, the proposed \modelshort\ can also help understanding tasks like image retrieval, by performing data augmentation using our \modelshort\ as a generator. We verified this idea by retraining a model that has state-of-the-art performance in image retrieval with the augmented data, and achieve significant improvement.





Our contributions can be summarized as follows:
\begin{itemize}
\item We introduce \modelshort, a new method of \modellong, and design three novel pre-training tasks that are especially effective for image-to-text generation.
\item We achieve state-of-the-art (SotA) results on COCO Captions, Flicker30k on all metrics, outperforming existing SotA and concurrent methods by a large margin. 
We also present extensive experiments and analysis to provide useful insights on the effectiveness of each pre-training task and model variant.
\item We employ \modelshort\ to help image retrieval, an understanding task, by performing data augmentation. After retraining, the model that has state-of-the-art performance still achieves significant improvement on all recall metrics.
\end{itemize}



\section{Related Work}
\label{sec:related}

\subsection{Pre-training for NLP Tasks}
Recently, pre-trained language models (LM) over large language corpus 
such as ELMo \cite{peters2018deep}, BERT~\cite{devlin2019bert}, 
GPT2~\cite{radford2019language}, and XLNet~\cite{yang2019xlnet} have shown great advances for NLP 
tasks. 
Among numerous works in natural language pre-training, we review three Transformer-based methods that are most relevant to our approach, namely MASS~\cite{song2019mass}, Unicoder~\cite{huang2019unicoder}, and BART~\cite{lewis2019bart}. 

MASS~\cite{song2019mass} adopts the encoder-decoder framework to predict masked fragments given the remaining part of the sentence. We also use the encoder-decoder framework and train our text-only model.
Unicoder is a universal language encoder pre-trained based on three pre-training tasks. The new tasks help the model learn mappings among different languages from more perspectives. 
BART~\cite{lewis2019bart} uses a denoising autoencoder for pre-training. Specifically, its pre-training objective is to reconstruct~\textit{the whole sentence}, which is substantially different from the masked language modeling in BERT. 
Our method is inspired by these works, but since images are not sequential data, we have to tailor our model for cross-modal tasks in particular. 



\subsection{Pre-training for Cross-modal Generation Tasks}
Very recently, several attempts have been made to pre-train models for cross-modal generation tasks.

Both VideoBERT~\cite{sun2019videobert} and CBT~\cite{sun2019cbt} are seeking to conduct pre-training for the video captioning task. But they perform pre-training only for the BERT-based encoder to learn bidirectional joint distributions over sequences of visual and linguistic tokens. So they have to train a separate video-to-text decoder. In contrast, Unified VLP~\cite{zhou2019unified}, concurrently with our work, uses a shared multi-layer transformer network for both encoding and decoding. Following UniLM~\cite{dong2019unified}, they pre-train the model on two masked language modeling (MLM) tasks, like cloze tasks designed for sequence-to-sequence LM. So target prediction is still masked tokens, not the whole sentence. However, we find that by using generative pre-training objectives such as \taskone\ , \tasktwo\ and \taskthree\ , \modelshort\ can outperform Unified VLP significantly on Image Captioning. 





\section{Preliminaries}
\label{sec:preliminiaries}
\textbf{Linguistic Representation.} For each token in the input language sequence, its representation is a sum of token embedding and position embedding. We denote the input tokens as $\wv=\{w_1, w_2,...,w_M\}$ and the corresponding representations as  $\xv^T=\{x^T_1, x^T_2,...,x^T_M\}$.\\ 
\textbf{Image Representation.} For each input image, we first detect objects using a pre-trained Faster R-CNN model \cite{anderson2018bottom}. Here, the top 100 objects with highest confidence scores are selected, each of which has a feature vector computed by mean-pooling the last-layer convolutional feature of its region of interest. 
To represent the position of each object, we construct a 5-d position vector from its spatial location (normalized top-left and bottom-right coordinates) and the fraction of image area it covered. 
Next, we concatenate the feature vector and position vector of each object and transform it into another vector by linear projection, to make sure the dimensions $h$ of linguistic tokens and visual tokens are identical, and we denote as image regions as $\vv=\{v_1, v_2,...,v_N\}$ and the corresponding representations as $\xv^I=\{x^I_1, x^I_2,...,x^I_N\}$.\\
\textbf{Image Refining.} Unlike words in text, image regions lack a natural ordering. To better model the relationship among objects in an image, we add an additional image refining layer following AoANet \cite{huang2019attention} to refine the image representation before feeding them to the encoder. We refer readers to Appendix~\ref{sec:aoa_refining_layer} for technical details.

\section{\modellong}
\subsection{Revisit Pre-training Tasks}
\label{sec:revisit}

In this section, we first review 
the objectives of the classical masked language modeling that is commonly used for understanding tasks.

The objective of masked language and region modeling is to learn joint representions for both vision and language by recontructing masked tokens ${\bar{\wv}}$ from a corrupted version $\hat{\wv}$:
\begin{equation}
\footnotesize
\max_{\theta} ~\Lcal_{MLM}=\log p_\theta(\bar{\wv}|\hat{\wv}) = \sum_{t=1}^{T} \log p_\theta (w_t | \hat{\wv})m_t
\end{equation}
where $m_t=1$ if $w_t$ is masked as corruption, and 0 otherwise. The objective is designed based on \emph{bidirectional} contexts which allows the words in the future to be attended. So only the masked tokens of the sentence or labels of regions are required to be predicted instead of the whole sentence.

The pros and cons of this pre-training objective can be concluded in the following aspects:
\begin{itemize}
    \item \emph{Downstream tasks}: Many BERT-based cross-modal pre-pretraining models choose masked language and region modeling as their main pre-training task. Because the encoder learned from this task can provide representations of both vision and language based on \emph{bidirectional} context and it is naturally fit for many understanding tasks, including VQA, Image Retrieval, etc.
    \item \emph{Architecture modification}: For downstream generation tasks, models pre-trained only through mask prediction tasks 
    usually have to train an additional layer for generation, as pointed out by Song et al.~\cite{song2019mass}. But separately trained decoders could create a pretrain-finetune discrepancy that hurt the generality of the model. This results in a gap between pre-training and fine-tuning on generation tasks. 
\end{itemize}
To address these concerns, we propose \modelshort\ and the new pre-training tasks. 
\subsection{Model Architecture}
\label{sec:framework}

\modelshort\ has a unified encoder and decoder architecture and can be pre-trained through different generative pre-training tasks. Basically, both encoder and decoder are multi-layer Transformer networks. The encoder reads the source image and sentence and generates a set of representations as introduced in Section~\ref{sec:preliminiaries}. Different from other BERT-based encoder-only models, the probability of each target token is estimated by the decoder given the \textit{cross-attention} performed over the final hidden layer of the encoder. 

Specially, we use shared parameters for encoder and decoder, and a faster attention strategy in decoder where the weights of self-attention and encoder-decoder attention are shared. We add a signal in the attention network to distinguish whether keys and values are from the output of the encoder for efficient re-use.
For our base model, we use 12 layers in the encoder and decoder. And we use 6 layers for our tiny model to explore the necessity of the decoder for image captioning in the experimental analysis (Section~\ref{sec:ablation}). 

\begin{figure}[t]
    \centering
    \begin{tabular}{p{0.5\textwidth} p{0.5\textwidth}}
  \vspace{0pt} \includegraphics[scale=0.25]{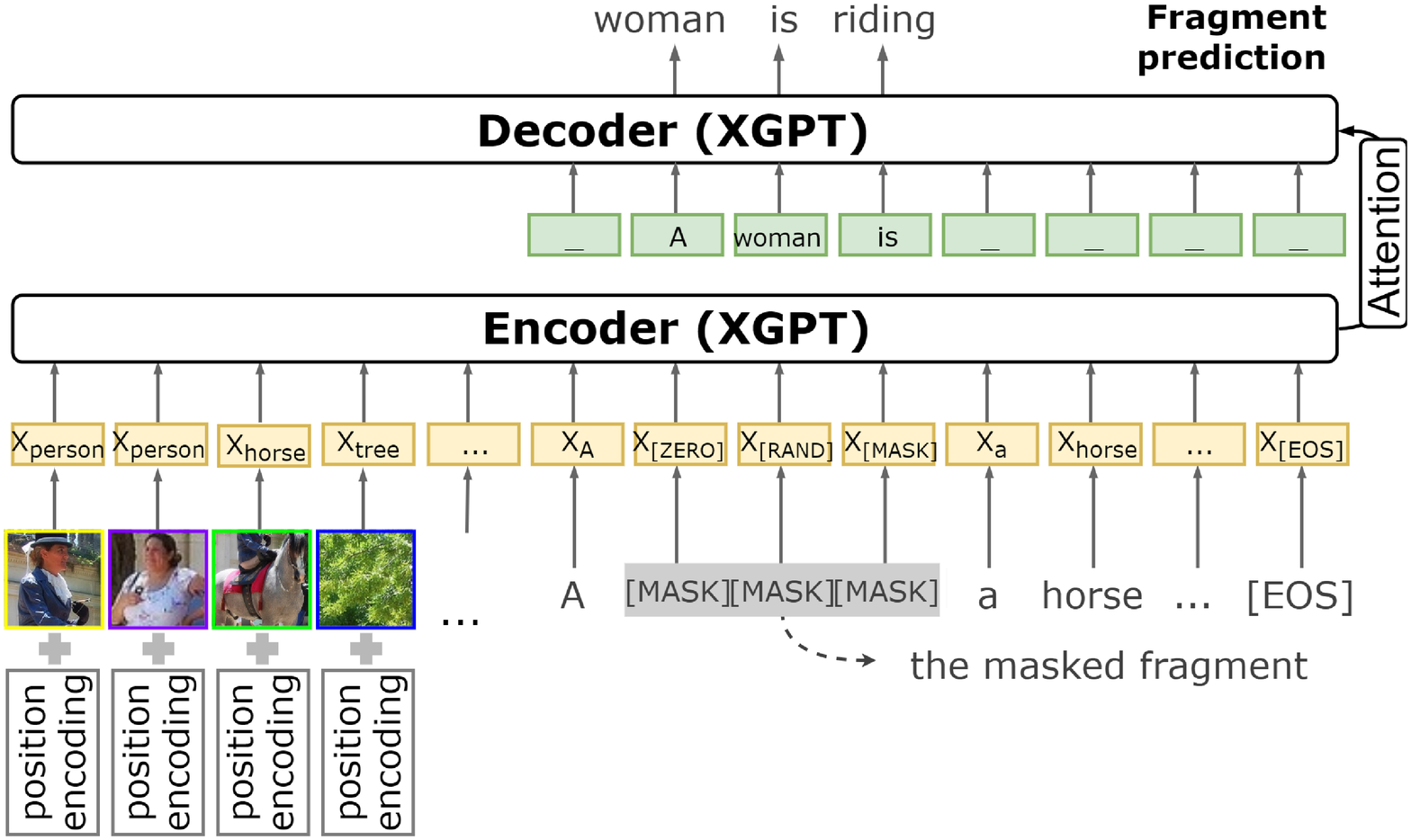} & \vspace{0pt} \includegraphics[scale=0.25]{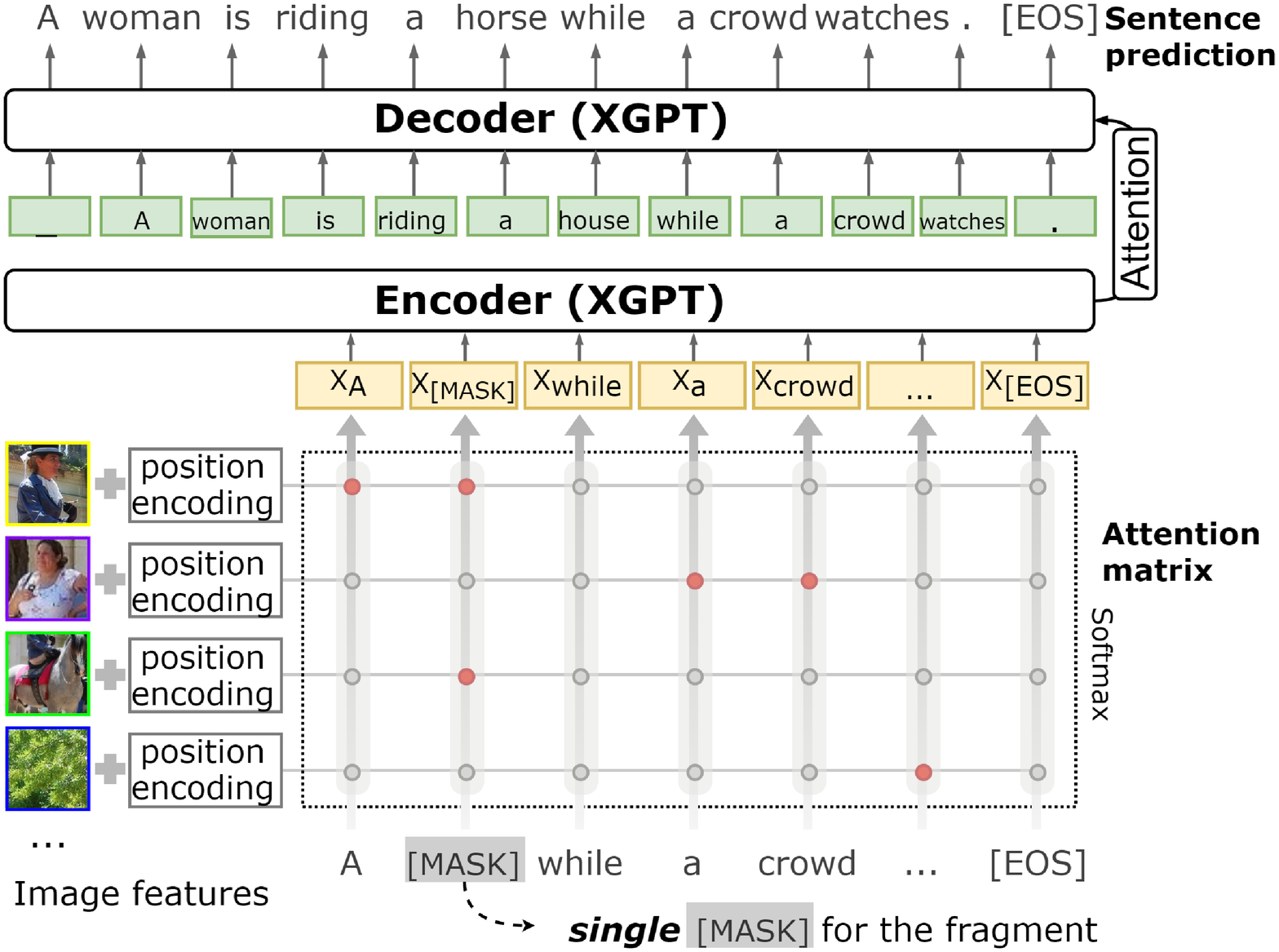}\\
    (a) \taskone\ & (b) \tasktwo\ \\
         \multicolumn{2}{c}{
         \includegraphics[scale=0.25]{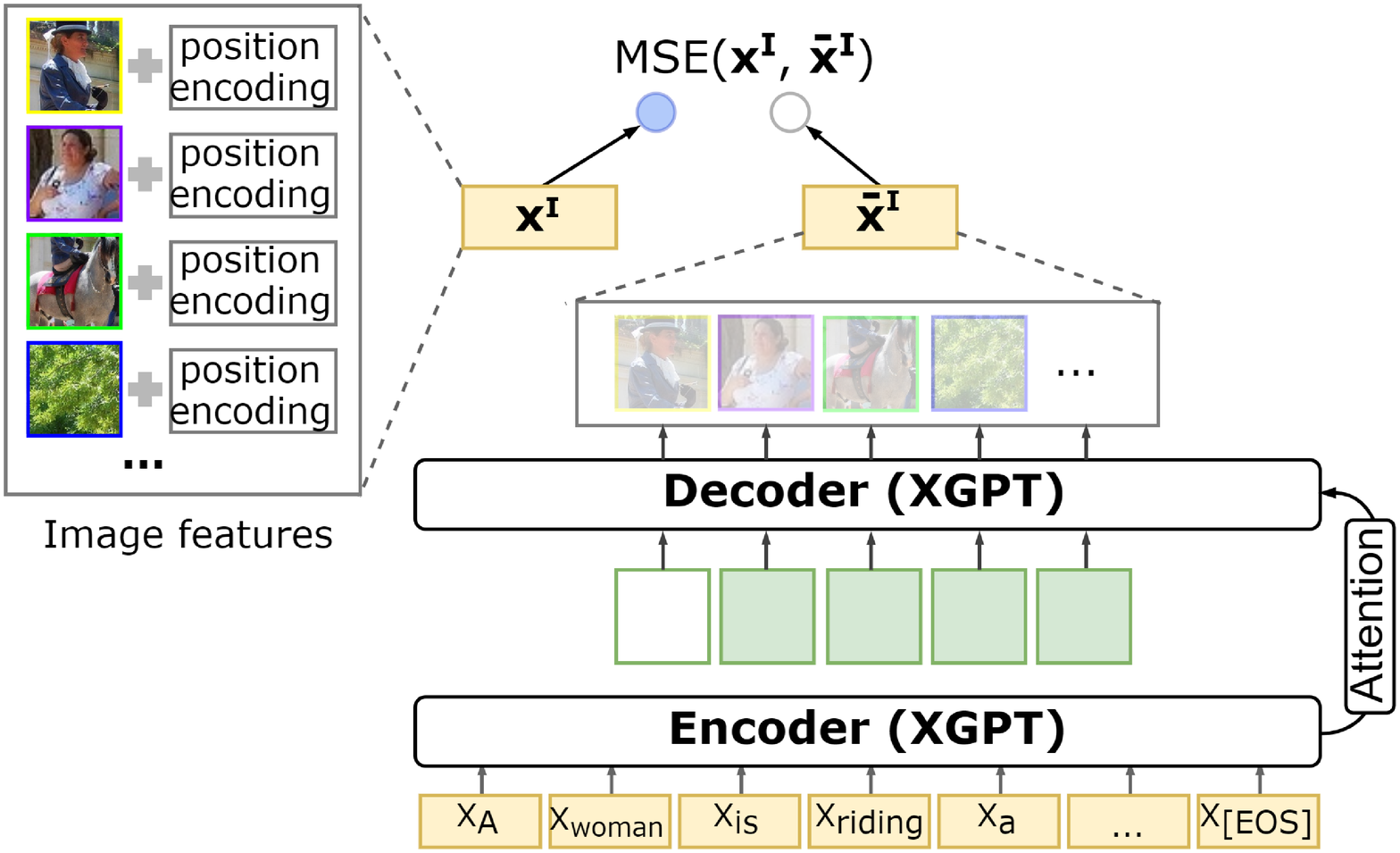}
         }\\
         \multicolumn{2}{c}{(c) \taskthree\ }
    \end{tabular}
    \caption{Cross-modal generative pre-training tasks}
    \label{fig:three_tasks}
\end{figure}

\subsection{Generative Pre-training Tasks}

Unlike Unified VLP, we use the image captioning task as a basic generative task in the pre-training stage. It only takes images as inputs (single modality). We also introduce three new cross-modal generative pre-training tasks that can jointly pre-train the encoder and decoder: \taskoneshort, \tasktwoshort, and \taskthreeshort. 
\\
\\
{\bf Image Captioning\ (IC).}\;\;\;\;
A major approach to Image Captioning is encoder-decoder framework. Giving the image regions $\vv$,  the training objective is to generate the caption ${\wv}$ in an autoregressive manner by minimize the negative log-likelihood
\vspace{-5pt}
\begin{equation}
\Lcal_{\text{IC}} = -\lambda_{\text{IC}} \sum_{t=1}^{T} \log p_\theta (w_t |\wv_{<t},\vv)
\end{equation} 
where $\wv_{<t}$ is the context history produced by the neural generator, and $\lambda_{\text{IC}}$ is the pre-defined weight of IC loss. This objective requires the model to predict the whole sentence from scratch.
\\
\\
{\bf \taskone\ (\taskoneshort).}\;\;\;\;
\taskoneshort\ aims to teach the model to learn the relationship between vision and language by predicting consecutive tokens in the decoder side.

\modelshort\ is trained to reconstruct the n-gram masked words through a sequence to sequence framework. This task is similar in the idea to n-gram MLM in BERT or Masked Seq-to-Seq in MASS. The difference lies in that (1) the task encourages the encoder to learn the cross-modality relationships between the unmasked tokens and image regions, and (2) the decoder has to generate masked tokens of the fragment, and extract useful image-conditioned information from the encoder side.


As shown in Figure~\ref{fig:three_tasks}(a), we concatenate the regions and unmasked tokens as input to the encoder during pre-training. And we let the decoder predict the masked fragment by minimzing the negative log-likelihood loss:
\vspace{-5pt}
\begin{equation}
\Lcal_{\text{\taskoneshort}} = -\lambda_{\text{\taskoneshort}}\sum_{t=1}^{M} \log p_\theta (w_t |\wv_{<t},\hat{\wv},\vv)m_t
\end{equation}
where $\hat{\wv}$ is the corrupted caption, and $m_t=1$ if $w_t$ is in the masked fragment, and $0$ otherwise. $\lambda_{\text{\taskoneshort}}$ is the pre-defined weight of \taskoneshort\ loss.
\\
\\
{\bf \tasktwo\ (\tasktwoshort).}\;\;\;\;
In \tasktwoshort\ we take distance between feature spaces of two modalities into account, and use an attention matrix to model the underlying text-image alignments. Besides, \tasktwoshort\ forces the model to reconstruct the whole sentence without giving it the length of the masked fragment, as illustrated in Figure~\ref{fig:three_tasks}(b). Specifically, we use

\begin{itemize}
    \item \textit{Single \texttt{[MASK]} token for fragments.} Inspired by text filling task in \cite{lewis2019bart}, we first sample n-gram fragments to be mask and, then, replace each with a single \texttt{[MASK]} token. This is more challenging because the model have to predict not only the missing tokens but also the length of the original sentence. We also compare this method with conventional method which keeps the placeholder for all masked tokens in Section~\ref{sec:ablation}.
    \item \textit{The attention-driven image-text matching.} To model the text-image alignments, we first compute an attention matrix for each token-region pair $(w_i, v_j)$: $A_{ij}=W [x_i^T, x_j^I, x_i^T\odot x_j^I]$ where $W\in \Rmat^{3\times h}$ is a trainable weight and $\odot$ is elementwise multiplication. Then, we represent each word as the weighted sum of all region representations based on the attention matrix: $x_i^T=\sum^N_{j=1}softmax(A_{ij})x_j^I$. Finally, \modelshort\ takes new $x_i^T$ as input and tries to predict the original word sequence $\wv$. The loss function is defined as
\vspace{-5pt}
\begin{equation}
\Lcal_{\text{\tasktwoshort}} = -\lambda_{\text{\tasktwoshort}}\sum_{t=1}^{M} \log p_\theta (w_t |\wv_{<t},\hat{\wv},\vv)
\end{equation}
where $\hat{\wv}$ is the corrupted caption, and $\lambda_{\text{\tasktwoshort}}$ is the pre-defined weight of \tasktwoshort\ loss.\\
\end{itemize}
{\bf \taskthree\ (\taskthreeshort).}\;\;\;\;
Text-to-image (T2I) generation can be regarded as the inverse problem of image captioning, a Image-to-Text (I2T) problem. It is natural and reasonable to unify the model to leverage the underlying semantic in both domains. 

In contrast to Uniter \cite{chen2019uniter}, which involved image feature generation by learning to regress the transformer output of each masked region to its visual features, TIFG aims to regress the decoder output of all image regions conditioned on text descriptions rather than only the masked regions. As shown in Figure~\ref{fig:three_tasks}(c), we employ the encoder-decoder pipeline to convert linguistic representations into $\bar{\xv}_i^{I}$ of the same length and dimension as image representations $\xv_i^{I}$. Then we train with mean squared error to supervise the \modelshort\ to generate semantically consistent image features. Mathematically, this loss can be expressed as:
\vspace{-1mm}
\begin{equation}
  \begin{aligned}
    \Lcal_{\text{\taskthreeshort}}=\lambda_{\text{\taskthreeshort}}\frac{1}{N}\sum_{i=1}^{N}\|\xv_i^{I}-\bar{\xv}_i^{I}\|^2_2
    \\
  \end{aligned}
\end{equation}
where $\lambda_{\text{\taskthreeshort}}$ is the weight of \taskthreeshort\ loss.
\\
\\
{\bf Multi-task pre-training.}\;\;\;\;Following Unicoder \cite{huang2019unicoder} 
, we calculate task-specific loss in turns and update the model for each task 
in every pre-training iteration. We include these tasks with individual weight of loss to study how each objective works for pre-training.

\section{Experiments and Results}
\label{sec:exp}

\subsection{Training Stages}

\textbf{Out-of-domain pre-training stage.}\;\;\;\;We first conduct pre-training on Conceptual Captions (CC) dataset \cite{sharma2018conceptual} which contains about 3M image-caption pairs scraped from alt-text enabled web images. The automatic collection leaves some noise (e.g., not relevant and too short) in the dataset but brings a massive scale. So we use it only as our out-of-domain dataset for the first pre-training stage. We set the weight to 1 for all pre-training tasks on CC.\\
\\
\textbf{In-domain pre-training stage.}\;\;\;\;Before fine-tuning \modelshort\ on the final image captioning task, we find it beneficial to further pre-train the model using the data from downstream tasks with the proposed pre-training objectives. This step allows the model to adapt to the target domain. So we reduce the weights of the cross-modal tasks (i.e., $\lambda_{\text{\taskoneshort}}, \lambda_{\text{\tasktwoshort}}, \lambda_{\text{\taskthreeshort}}$) and keep the image caption task unchanged.\\
\\
\textbf{Fine-tuning stage.}\;\;\;\;In this step, the model only takes image features and position information as input, and the decoder is trained to predict the whole sentence in an autoregressive manner. We also applied other inference approaches like beam search as well. More details follow Section~\ref{sec:implementation}.

\subsection{Evaluation datasets} The datasets for downstream tasks include COCO Captions \cite{chen2015microsoft} and Flickr30k \cite{young2014flick}. In these datasets, each image is labeled with 5 captions. We follow Karpathy's split\footnote{cs.stanford.edu/people/karpathy/deepimagesent/caption\_datasets.zip}, which gives 113.2k/5k/5k and 29.8k/1k/1k images for train/val/test splits respectively. 
We use standard metrics for Image Captioning, including \texttt{BLEU@4}, \texttt{METEOR}, \texttt{CIDEr}, \texttt{SPICE}, to evaluate the propose method and compare with other methods.

\subsection{Implementation Details}
\label{sec:implementation}
In all experiments, the backbone Transformer of \modelshort\ follows Vaswani et al.~\cite{vaswani2017attention}, and we modify it to the 
BERT-based encoder-decoder architecture with 768 hidden units, 8 heads, GLEU activations used as GPT \cite{hendrycks2016bridging}. The dropout rate is 0.1.

We train \modelshort\ with mixed-precision training and FP16, which makes use of GPUs more efficiently. The Adam \cite{kingma2015adam} 
with $\beta_1=0.9$, $\beta_2=0.98$ is used for optimization.
The learning rate is varying from $1e-4$ to $2e-5$ for out-of-domain pre-training with invert square root decay \cite{vaswani2017attention}. The weight of the individual loss is set to 1.
During the in-domain pre-training and fine-tuning stage, we take an average of the top-4 pre-trained weights and reduce the initial learning rate to $1e-5$. 
We include $\lambda_{\text{\taskoneshort}}, \lambda_{\text{\tasktwoshort}}, \lambda_{\text{\taskthreeshort}}$ with 0.3 and $\lambda_{\text{IC}}=1$ for in-domain pre-training, and turn off other tasks during fine-tuning (image captioning only). 
The two-stage pre-training takes about 8 days to converge on 8x V100 GPUs with a total batch size of 512 by gradient accumulation.
We fine-tuned the model 30 epochs with four GPUs. 

For caption inference, we use
greedy search on the validation set and beam search with
beam size 2 on the test set. We describe more training details in Appendix~\ref{sec:implementation_details}. 



%

\subsection{Experimental Settings}
\label{sec:exp_settings}

We compare \modelshort\ with state-of-the-art methods on image captioning in two settings:
\begin{itemize}
    \item \lpretrain\ is our baseline setting where \modelshort\ is trained from scratch with two pre-trained objectives: Masked Language Model in BERT \cite{devlin2019bert} and Masked Seq-to-Seq in MASS \cite{song2019mass}, and directly fine-tuned on the image-caption task.
    \item \vlpretrain. Weights are also initialized from Text Pre-training. To fuse the information from Image-Text during the pre-training stage, \modelshort\ continue pre-training on three proposed tasks along with the image captioning task, and fine-tuned only on the image captioning task in the end.
\end{itemize}

\subsection{Comparisons against SotAs}
Results comparing our methods and SotA methods on the test set are in \ref{tab:main_result}. We include state-of-the-art works (mostly without pre-training) and a recent work \cite{zhou2019unified} that also use pre-trained models, and our methods (the lowest part). All methods use a single model for the image captioning for a fair comparison. Our full model (Base) significantly outperforms SotA methods on all four metrics on both COCO Captions and Flickr30k.\\ 
\\
\textbf{Compare with baseline.}\;\;\;\;Compared to our baseline model that only uses text pre-training, cross-modal pre-training tasks improves the performance on all metrics, which validates the importance of \vlpretrain\ for generation tasks. Unified VLP also employes two masked language pre-training tasks and provides a baseline initialized from a language pre-trained model (UniLM)\cite{dong2019unified}. The improvement is significant on both benchmark datasets, and is particular sound on Flickr30k, including absolute 17.9\% gain on \texttt{CIDEr}, 6.1\% on \texttt{BLEU@4}, 3.3\% on \texttt{METEOR}, and 2.9\% on \texttt{SPICE}.  
Comparing the gain from \vlpretrain\ tasks, \modelshort\ achieves higher improvement than Unified VLP, demonstrating the effectiveness of the three designs in \modelshort.

\begin{table*}[t!]
\centering
\begin{tabular}{l@{\hspace{18pt}}cccc@{\hspace{18pt}}cccc}
\toprule
\multirow{2}{*}{\begin{tabular}[c]{@{}c@{}}Model\end{tabular}} & \multicolumn{4}{c}{Flick30k} & \multicolumn{4}{c}{COCO}\\ 
& \texttt{C} & \texttt{B@4} & \texttt{M} & \texttt{S} & \texttt{C} & \texttt{B@4} & \texttt{M} & \texttt{S} \\ 
\midrule
\multicolumn{9}{l}{Approaches that \textbf{\emph{do not}} use any pre-trained model}\\
BUTD~\cite{anderson2018bottom} & 56.6 & 27.3 & 21.7 & 16.0 & 113.5 & 36.2 & 27.0 & 20.3 \\
NBT (with BBox)~\cite{lu2018neural} & 57.5 & 27.1 & 21.7 & 15.6 & 107.2 & 34.7 & 27.1 & 20.1 \\
GCN-LSTM (spa)~\cite{yao2018exploring} & - & - & - & - & 115.6 & 36.5 & 27.8 & 20.9\\ 
GCN-LSTM (sem)~\cite{yao2018exploring} & - & - & - & - & 116.3 & 36.8 & 27.9 & 20.9 \\
GVD~\cite{zhou2019grounded} & 62.3 & 27.3 & 22.5 & 16.5 & - & - & - & - \\
AoANet~\cite{huang2019attention} & - & - & - & - & 119.8 & 37.2 & 28.4 & 21.3 \\
\midrule
\multicolumn{9}{l}{Approaches that \textbf{\emph{do}} use pre-trained models or pre-trained language models}\\
$\text{Unified VLP}^\star$~\cite{zhou2019unified} & 56.8 & 27.6 & 20.9 & 15.3 & 114.3 & 35.5 & 28.2 & 21.0 \\
$\text{Unified VLP}$~\cite{zhou2019unified} & 67.4 & 30.1 & 23.0 & 17.0 & 116.9 & 36.5 & 28.4 & 21.2 \\
${\textbf{\modelshort}}^\star$ & 53.0 & 25.7 & 20.3 & 14.7 & 113.0 & 34.4 & 27.8 & 20.8 \\
${\textbf{\modelshort}}$ & \textcolor{blue}{\textbf{70.9}} & \textcolor{blue}{\textbf{31.8}} & \textcolor{blue}{\textbf{23.6}} & \textcolor{blue}{\textbf{17.6}} & \textcolor{blue}{\textbf{120.1}} & \textcolor{blue}{\textbf{37.2}} & \textcolor{blue}{\textbf{28.6}} & \textcolor{blue}{\textbf{21.8}} \\
\bottomrule
\end{tabular}
\caption{Comparison with the previous state-of-the-art methods. \textcolor{blue}{\textbf{Bold}} indicates best value overall. $\text{Unified VLP}^\star$ and ${\text{\modelshort}}^\star$ perform \lpretrain. The former is initialized from UniLM, while the latter is pre-trained from scratch with less text data, which is detailed in Section~\ref{sec:exp_settings}. Both Unified VLP and \modelshort\ perform \vlpretrain\  (see Section~\ref{sec:exp_settings}) where the weights are initialized from Text Pre-training and pre-trained on different tasks, respectively.}
\label{tab:main_result}
\end{table*}

\begin{table*}[t!]
\centering
\begin{tabular}{llcccc}
\toprule
\multirow{2}{*}{\begin{tabular}[c]{@{}c@{}}Stage\end{tabular}}  & \multirow{2}{*}{\begin{tabular}[c]{@{}c@{}}Pre-training Tasks\end{tabular}} & \multicolumn{4}{c}{COCO} \\
 &               & \texttt{C} & \texttt{B@4} & \texttt{M} & \texttt{S} \\
\midrule
\multirow{5}{*}{\begin{tabular}[c]{@{}c@{}}Out-of-domain (CC)\end{tabular}} 
& IC & 116.4 & 35.9 & 28.2 & 21.1 \\
& IC + \taskoneshort & 117.7 & 36.2 & 28.2 & 21.2 \\
& IC + \tasktwoshort & 118.1 & 36.4 & 28.3 & 21.3 \\
& IC + \taskthreeshort & 117.3 & 36.0 & 28.2 & 21.2 \\
& IC + \taskoneshort\ + \tasktwoshort\ + \taskthreeshort & \textbf{118.3} & \textbf{36.5} & \textbf{28.4} & \textbf{21.3}\\
\midrule
\multirow{4}{*}{\begin{tabular}[c]{@{}c@{}}Out-of-domain (CC) \\
+ In-domain (COCO)\end{tabular}} 
& IC + \taskoneshort & 119.1 & 36.7 & 28.5 & 21.5 \\
& IC + \tasktwoshort & 119.2 & 36.6 & 28.5 & 21.6 \\
& IC + \taskthreeshort & 118.2 & 36.4 & 28.4 & 21.3 \\
& IC + \taskoneshort\ + \tasktwoshort\ + \taskthreeshort & \textbf{120.1} & \textbf{37.2} & \textbf{28.6} & \textbf{21.8}\\
\bottomrule
\end{tabular}
\caption{Ablation analysis of pre-training tasks on COCO Captions.}
\label{tab:task_comparison}
\end{table*}

\begin{table}[t!]
\small
\centering
\begin{tabular}{lcccc}
\toprule
Model & \texttt{C} & \texttt{B@4} & \texttt{M} & \texttt{S} \\ 
\midrule
Tiny \enconly & 112.1 & 34.1 & 27.9 & 20.6  \\
Tiny \encdec & 110.8 & 33.7 & 27.6 & 20.5  \\
Tiny \encdecshare & \textbf{112.7} & \textbf{34.6} & \textbf{27.9} & \textbf{20.7} \\
Base \encdecshare & \textbf{112.9} & \textbf{35.0} & \textbf{27.9} & \textbf{20.8} \\
\bottomrule
\end{tabular}
\vspace{2mm}
\caption{Evaluation results on COCO Captions using different model structures. We use 6-layer Transformers for Tiny models, and 12-layer for Base models and directly train the model on image captioning task without any pre-training.}
\label{tab:structure_comparison}
\end{table}

\begin{table}[t!]
\small
\centering
\begin{tabular}{lcccc}
\toprule
 & \texttt{C} & \texttt{B@4} & \texttt{M} & \texttt{S} \\
\midrule
multi \texttt{[MASK]} & 117.8 & 36.1 & 28.2 & 21.3 \\
single \texttt{[MASK]} & \textbf{118.1} & \textbf{36.4} & \textbf{28.3} & \textbf{21.3} \\
\bottomrule
\end{tabular}
\vspace{2mm}
\caption{Comparison of two masking methods on COCO Captions.}
\label{tab:mask_comparison}
\end{table}
\begin{table}[t!]
\small
\centering
\begin{tabular}{llll}
\toprule
 & \texttt{R@1} & \texttt{R@5} & \texttt{R@10} \\
\midrule
ViLBERT \cite{lu2019vilbert} & 58.2 & 84.9 & 91.5 \\
ViLBERT + augmentation & \textbf{60.4} & \textbf{86.4} & \textbf{91.9} \\
\bottomrule
\end{tabular}
\vspace{2mm}
\caption{Results of image retrieval task on Flickr30k.}
\label{tab:image_retrieval}
\end{table}
\section{Analysis}
\label{sec:ablation}

\textbf{Is Decoder Necessary?}\;\;\;\;
To find the best model structure for image captioning, we also designed three model variants.
\textbf{\enconly} is a single multi-layer Transformer that takes either image region features for the undirectional model or a pair of text and image packed together, uses different
self-attention masks to control the access to the context, which is similar to UniLM and Unified VLP. \textbf{\encdec} is a Transformer encoder-decoder architecture in which all the weights are initialized randomly and not shared between the encoder and decoder. \textbf{\encdecshare} is like \encdec, but the parameters between encoder and decoder are shared. We also use a signal in the attention network to control whether keys and values are from the encoder. This greatly reduces the memory footprint of the
model.

In all settings, models are trained on the image captioning task without any text or image pre-training. 

Table~\ref{tab:structure_comparison} reports results of these settings on two model sizes: Base (layers=12), Tiny (layers=6). 
With a similar number of parameters, the shared setup with 6-layer (Tiny \encdecshare) performs better than the encoder and decoder parameters are not shared. Tiny \enconly\ model which simply reuses the encoder for decoding can outperform Tiny \encdec, although it performs less well than EncDecShare models. 
We also notice that the model with shared 12-layer encoder and decoder parameters (Base \encdecshare) performs best. We use this as the optimal architecture that achieves the new state-of-the-art results in Table~\ref{tab:main_result}. \\
\\
\textbf{Effectiveness of Proposed Tasks.}\;\;\;\;
We analyze the effectiveness of different pre-training tasks through ablation studies over COCO Captions and Flickr30K. The results are shown in Table~\ref{tab:task_comparison}. Firstly, we establish our baseline: Row~1 in Table~\ref{tab:task_comparison} shows the results of the model pre-trained on out-of-domain datasets through the image captioning task only. 

As for the out-of-domain pre-training stage, there are significant improvements across all three tasks (comparing Row~2,3,4 with Row~1 baseline). Among the three, we observe \tasktwoshort\ which helps the model to learn text-image alignments achieves the biggest improvement, while \taskthreeshort\ the smallest. This is probably because of the discrepancy of the decoder which is originally designed to predict captions and the task objective which is to predict all image region features. When combining all three tasks, we find that they are complementary to each other and see the highest gain of approximately +1.9 on \texttt{CIDEr} over Row~1. 

Comparing with one-stage pre-training, we find that each task combination pre-trained after the second stage (Row~6-8) gains approximately +2 on \texttt{CIDEr}. This indicates that two-stage pre-training with in-domain data enables the model to adapt to the downstream data better than only using out-of-domain pre-training. Combining all three tasks leads to the highest score on all metrics. We use this as the optimal pre-training setting for further experiments.\\
\\
\textbf{How to Efficiently Mask?}\;\;\;\;
We conduct further experiments to compare two masking strategies for \tasktwoshort: (1) Multi \texttt{[MASK]}: replace tokens in the sampled fragment with exactly the same number of \texttt{[MASK]} tokens, (2) Single \texttt{[MASK]}: replace the fragment with a single \texttt{[MASK]} token. Results shown in Table~\ref{tab:mask_comparison} suggest that the effectiveness of IDA highly dependent on the masking strategy. Single \texttt{[MASK]} is obviously better than Multi \texttt{[MASK]}. Multi \texttt{[MASK]} provides the decoder with full position information of masked token, which reduced the difficulty for the decoder to predict correct words, thus, gives a less well performance (-0.3 on \texttt{CIDEr}). Therefore, we use the single \texttt{[MASK]} as the optimal pre-training setting. 

\subsection{Data Augmentation for Image Retrieval.}\;\;\;\;
In addition to generation tasks, our \modelshort\ can also help vision-and-language understanding tasks, such as image retrieval, by performing data augmentation as an image description generator. Image retrieval is a task of identifying an image from a pool given a caption describing its content. We generate 62k more captions for all 29k images (about 2.1 captions for each) in the Flickr30k training set, which originally contains 145k captions. We continue to fine-tune the open-source state-of-the-art model\footnote{https://github.com/jiasenlu/vilbert\_beta} introduced in \cite{lu2019vilbert} on the combination of the augmentation and the original training data.

Comparing Row~1 (trained only with the original training data) against Row 2 (fine-tuned on augmented data) in  Table~\ref{tab:image_retrieval}, 
the improvement is significant ($2.2\%$ on \texttt{R@1}, $1.5\%$ on \texttt{R@5}, and $0.4\%$ on \texttt{R@10}). The higher relative gain on \texttt{R@1} also indicates that the generator can produce high-quality image captions which can help the model better understand images. 

A negative example of the generation results is provided in Table~\ref{tab:negative_example}. The first sentence contains wrong information (brown$\xrightarrow{}$black); the second has a duplicated phrase. Both can be considered as noise.

Table~\ref{tab:generation_sample} shows a positive example of \modelshort-generated captions. We can see that the generated captions are grammatically and semantically correct, and also can increase the diversity of the data. 

\begin{table*}[t!]
\small
\fontdimen2\font=0.5ex
\centering
\begin{tabular}{@{\hspace{-5pt}}l@{\hspace{-1pt}}l}
\toprule
\multirow{10}{*}{\
\begin{tabular}[c]{c}\includegraphics[trim={3cm 0cm 5cm 0cm},clip,width=2.3cm]{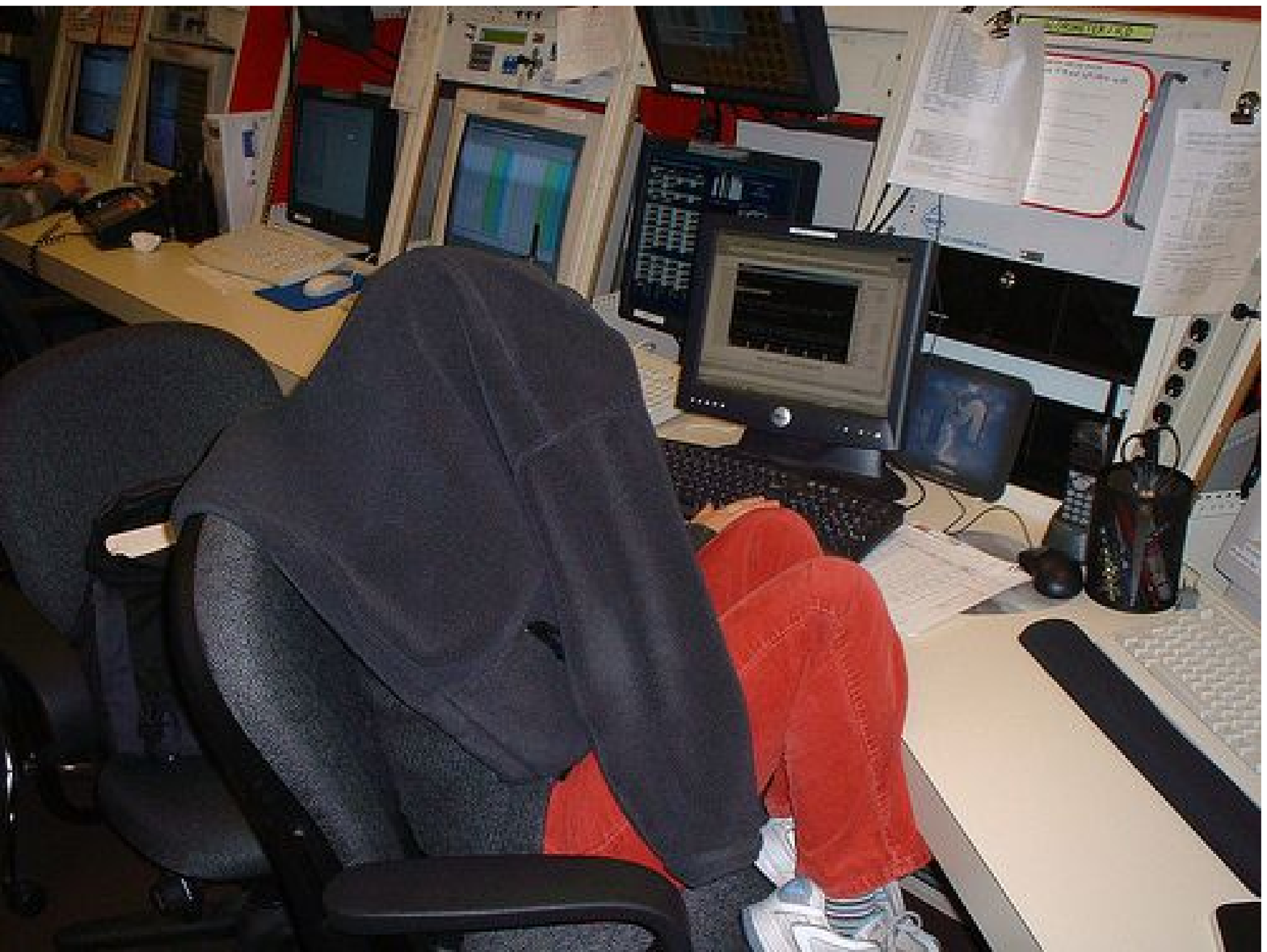}\end{tabular}}
& \textbf{ Human-generated captions}\\
& A person trying desperately \underline{\smash{not to be photographed}} by putting their sweater over ...\\
& A person wearing red pants \underline{\smash{hides their head}} under a \underline{\smash{black jacket}} in front of a desk ...\\
& A person with red pants with cover over her head \underline{\smash{sitting}} in front of multiple computers.\\
& \underline{\smash{The woman}} tries to \underline{\smash{hide from work}} under a black sweatshirt, but her red corduroy ...\\
& \underline{\smash{a child}} is \underline{\smash{hiding}} under a \underline{\smash{sweater}} in a chair.\\
\cmidrule{2-2}
& \textbf{ \modelshort-generated captions}\\
& \underline{\smash{A woman}} wearing red pants is \underline{\smash{sitting}} at a desk in front of a computer.\\
& A person in a \underline{\smash{blue sweatshirt}} is \underline{\smash{sleeping}} in a chair.\\
& A woman in a \underline{\smash{blue sweatshirt}} is \underline{\smash{sleeping}} in a chair in front of a computer.\\
\bottomrule
\end{tabular}
\caption{An example of generated captions for the given image. \underline{\smash{Underlined}} text shows the difference between captions. We can see that in the original training data, underlined text is usually people's guess with personal emotions, e.g., \textit{hide from work}. While the generated captions provide more modifier variants (e.g., \textit{blue}) and verb variants (e.g., \textit{sleeping}) according to what can be seen in the picture. }
\label{tab:generation_sample}
\end{table*}

\begin{table*}[t!]
\small
\fontdimen2\font=0.5ex
\centering
\begin{tabular}{@{\hspace{-5pt}}l@{\hspace{-1pt}}l}
\toprule
\multirow{9}{*}{\
\begin{tabular}[c]{c}\includegraphics[trim={0cm 0.5cm 0cm 0cm},clip,width=2.3cm]{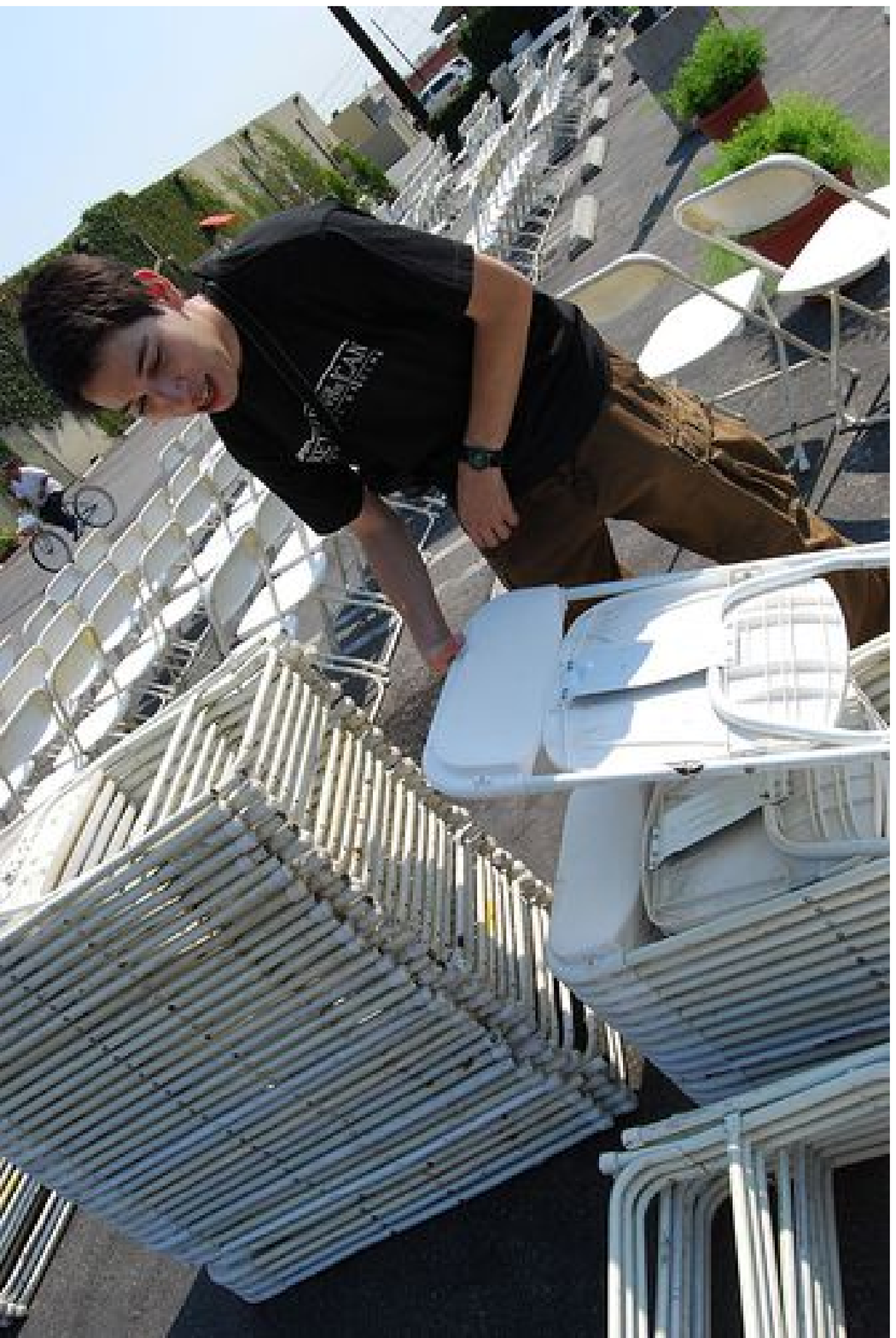}\end{tabular}}
& \textbf{ Human-generated captions}\\
& A young boy wearing a black shirt, BROWN pants and a black watch has his hand on a ...\\
& A young man wearing a black shirt takes a folding chair from a large stack.\\
& A person helping to set up chairs for a big event.\\
& A young man in a black shirt stacks chairs.\\
& A boy is setting up folding chairs.\\
\cmidrule{2-2}
& \textbf{ \modelshort-generated captions}\\
& A man in a black t shirt and \underline{\smash{black shorts}} is putting up a white chair.\\
& A man in a black t shirt and \underline{\smash{black t shirt}} works on a folding chair.\\
\bottomrule
\end{tabular}
\caption{A nagative example of the generation results. The first predicted the wrong color of the pants (brown$\xrightarrow{}$black). And the second generated caption duplicate the same phrase (black t shirt).}
\label{tab:negative_example}
\end{table*}

\section{Conclusion}
\label{sec:conclusion}
In this paper, we present \modelshort, \modellong. Three main pre-training tasks are proposed and the ablation study shows that the effectiveness of each task is different. The combination of all tasks achieves stronger performance on all evaluation metrics suggested that they are complementary to each other. After in-domain and out-of-domain pre-training, \modelshort\ outperforms state-of-the-art models by a significant margin. For future works, we are curious about extending \modelshort\ to cross-modal understanding tasks, such as VQA and VCR.

\clearpage
%
%
\bibliographystyle{unsrt}
\bibliography{arxiv}

\newpage
\appendix
\section{Appendices}
\label{sec:appendix}

\subsection{AoA refining layer}
\label{sec:aoa_refining_layer}
Instead of directly feeding image region features to the encoder, we build a refining network to refine their representation using an AoA module following \cite{huang2019attention}. The AoA module adopts the multi-head attention function \cite{vaswani2017attention} where $\Qmat=W_Q\xv$, $\Kmat=W_k\xv$, and $\Vmat=W_v\xv$ are three individual linear projections of the region features $\xv$. The AoA layer is formulated as
\begin{equation*}
    \footnotesize
    \begin{aligned}
    AoA(f_{att}, &\Qmat, \Kmat, \Vmat) = \sigma(W_1\Qmat
    +W_2f_{att}(\Qmat, \Kmat, \Vmat)+b_1)\\
    &\odot(W_3\Qmat+W_4f_att(\Qmat, \Kmat, \Vmat)+b2)
    \end{aligned}
\end{equation*}
where $W_1,W_2,W_3,W_4\in \RR^{h\times h}, b_1,b_2\in\RR^{h}$ are learnable weights, $f_{att}$ is a conventional attention module which operates on some queries, keys and values and generates some weighted average vectors.

The refined region features $\xv'$ are calculated by
\begin{equation*}
    \begin{aligned}
    \footnotesize
    \xv' = LayerNorm(&\xv+AoA(f_{att}, W_q\xv, W_k\xv, W_v\xv))
    \end{aligned}
\end{equation*}

\subsection{More Implementation Details}
\label{sec:implementation_details}

For tokenization, we follow the line of ~\cite{koehn2007moses} to build vocabulary from English Wikipedia, and use byte-pair encoding (BPE) to process image captions. And we trim the max sequence length to 60 and represent each input image as 100 object regions. 

Following \cite{devlin2019bert}, the masked tokens in the encoder will be a \texttt{[MASK]} token 80\% of the time, a random token 10\% of the time and an unchanged token 10\% of the time. For \taskoneshort, we set the fragment length as roughly 50\% of the total number of tokens in the sentence. And the span lengths are drawn from a Poisson distribution ($\lambda=3$) for \tasktwoshort. Each span is replaced with a single \texttt{[MASK]}.

\end{document}